\documentclass{article} % For LaTeX2e
\usepackage{nips14submit_e,times}
\usepackage{hyperref}
\usepackage{url}
\usepackage[square,numbers]{natbib}
\usepackage[normalem]{ulem}
\usepackage{graphicx}
\usepackage{amsfonts}
\usepackage{amsmath,amssymb}
\usepackage{multirow}
\usepackage{array}
\usepackage{caption}
\usepackage{algpseudocode}
\usepackage{enumitem}
\algnotext{EndFor}
\algnotext{EndIf}

\usepackage{listings}
\usepackage{balance}

\lstnewenvironment{code}[2]{\lstset{caption=#1,label=#2}}{}
\newenvironment{inline}[1]{{\small\ttfamily #1}}{}

\title{Learning Probabilistic Programs}

\author{
Yura Perov\textsuperscript{1, 2} \\
\texttt{perov@robots.ox.ac.uk} \\
\And
Frank Wood\textsuperscript{1} \\
\texttt{fwood@robots.ox.ac.uk} \\
}

\nipsfinalcopy % Uncomment for camera-ready version

\begin{document}

\maketitle

\vspace{-7mm}
\begin{center}
{\small \textsuperscript{1}Department of Engineering Science, University of Oxford, Oxford, United Kingdom \\
\textsuperscript{2}Institute of Mathematics and Computer Science, Siberian Federal University, Krasnoyarsk, Russia}
\end{center}
\vspace{7mm}

\begin{abstract}
% !TEX root =  paper.tex

We develop a technique for generalising from data in which models are samplers represented as program text.
We establish encouraging empirical results that suggest that Markov chain Monte Carlo probabilistic programming inference techniques coupled with higher-order probabilistic programming languages are now sufficiently powerful to enable successful inference of this kind in nontrivial domains.
We also introduce a new notion of probabilistic program compilation and show how the same machinery might be used in the future to compile probabilistic programs for efficient reusable predictive inference.
\end{abstract}

\section{Introduction}
\label{sec:introduction}
% !TEX root =  paper.tex

In the context of Turing-complete, higher-order probabilistic programming languages \cite{wood2014anglican,goodman2008church,venture}, a probabilistic program is simultaneously a generative model and a procedure for sampling from the same.
All probabilistic programming procedures are program text that describe how to generate a sample value conditioned on the value of arguments.   A probabilistic programming procedure is a constructivist description of a conditional distribution.   Deterministic procedures merely encode particularly simple, degenerate conditional distributions. 

Higher-order probabilistic programming languages open up the possibility of doing inference over generative model program text directly via a generative prior over program text and the higher-order functionality of \inline{eval}. This paper is a first step towards the ambitious goal of inferring generative model program text directly from example data.    Inference in the space of program text is hard so, as a start, we present an account of our effort to directly infer sampler program text that, when evaluated repeatedly, produces samples with similar summary statistics to observational data.  

There are reasons to make this specific effort itself.  One is the potential automation of the development of new entries in the special collection of efficient sampling procedures that humankind has painstakingly developed over many decades for common  distributions, for example the Marsaglia \cite{marsaglia1964convenient} and Box-Muller \cite{box1958note} samplers for the normal distribution (see \cite{Devroye86non-uniformrandom} for others). 
 In this paper we develop preliminary evidence that suggests that such automated discovery might indeed be possible.  In particular we perform successful leave-one-out experiments in which we are able to learn a sampling procedure for one distribution, i.e. Bernoulli, given only program text for others and observed samples.  We do this by imposing a hierarchical generative model of sampling procedure text, fitting it to out-of-sample, human-written sampler program text, then inferring the program text for the left-out random variate distribution type given only sample values drawn from the same.

The second reason for making such an effort has to do with ``compiling'' probabilistic programs.  What we mean by compilation of probabilistic programs is somewhat more broad than both transformational compilation \cite{wingate2011lightweight} which compiles a probabilistic program into an MH sampler for the same and normal compilation of a probabilistic program to machine code that encodes a parallel forward inference algorithm \cite{Paige-ICML-2014}.  
What we mean by probabilistic program compilation is the automatic generation of program text that when run will generate samples distributed ideally identically to the posterior distribution of quantities of interest in the original program, conditioned on the observed data.  Concisely; given samples resulting from posterior inference in a probabilistic program, our aim is to learn program text that when evaluated generate samples from the same directly.  The reason for expressing and approaching compilation in this generality is that simpler approaches to generalizing probabilistic programming posterior samples via a less-expressive model families will suffer precisely due to the compromise in expressivity.  Distributions over expressions are valid posterior marginals in higher-order probabilistic programming languages.   Compiled probabilistic programs must be capable of generating the same.  This effort is also a first step towards such a compiler.

\section{Related Work}
\label{sec:relatedwork}
% !TEX root =  paper.tex

%``Background and Motivation''

Our approach to learning probabilistic programs relates to both program induction and statistical generalization from sampled observations.  The former is usually treated as search in the space of program text where the objective is to find a deterministic function that exactly matches outputs given parameters.  The latter,  generalizing from data, is usually referred to as either density estimation or learning.   %With respect to this, recent work on automatic modeling is the most similar to what we propose in this paper.

%\subsection{Program Induction}
\subsection{Automatic programming}

An extensive review and introduction is given in \cite{gulwani_et_al:DR:2014:4507} and its references.
Modern examples of functional inductive programming include IGOR2 \citep{kitzelmann2009analytical} in which search is utilized to find programs that match constraints specified by equations and MagicHaskeller \citep{katayama2011magichaskeller} which uses traditional search and brute force enumeration to find programs that obey constraints specified in terms of parameter/output pairs.
Similar search procedures are used to find constraint satisfying hypotheses in inductive logic programming \cite{muggleton1992golem,muggleton1994inductive} and probabilistic variants thereof \cite{muggleton1996stochastic,kersting2005inductive,de2008probabilistic}. Alternative search techniques such as genetic programming have also been used to find constraint satisfying programs,
and in some of this work it has been suggested that search is easier in the space of functional programming languages than in imperative \citep{briggs2006functional}.

%Deterministic program induction and automatic modelling includes work in evolutionary programming (genetic programming) \cite{genetic programming}, inductive programming (both logic \cite{logicinductiveprogramming} and functional \cite{notsurewhattocallthis}), and program synthesis \cite{}.  \comment{Yura -- fix this -- The general problem has been formulated in different ways, for example through provided train set of input-output pairs or formal specifications in different forms, which should be transformed to the desired efficient program or model. [Cite here Inductive Logic Programming, Dagstual Seminar, Inductive Stochastic Logic Programming, Sumit Gulvani, ... James Lloyd, Grosse.]  Different approaches are used and combined, including search, optimization, inference, and enumeration with heuristics.}

This insight supports an interesting choice made by \citet{liang2010learning} in searching over the  space of combinatory logic rather than lambda calculus expressions.   Our work is framed similarly to theirs (and the theoretical suppositions in \cite{mansinghka2009natively}) in that we impose a prior on the program text and use Bayesian inference machinery to infer a distribution over program text given observations.  Unlike \cite{liang2010learning} we learn stochastic programs from sampled observation data rather than deterministic programs from input/output pairs. 

%Bayesian approaches to program induction start with on prior over programs (cite Percy), written in functional language with explanable dependencies (cite Briggs and Neill) and lack of mutation in the state, were previously introduced and advocated. (Cite Daniel Tarlow here? Or at least in the discussion surely!)

\subsection{Generalizing from Data and Automated Modelling}

Generalizing from data is one of the main objectives of the fields of machine learning and statistics.  It is important to note that what we are doing here is a substantial departure from almost all prior art in these fields in the sense that the learned representation of the observed data is that of generative sampling program text rather than, say, a parametric or nonparametric model from which samples can be drawn using some extrinsic algorithm.   In our work the model is the sampler itself and it is represented as program code.

The greedy search over generative models structures in \cite{grosse2012exploiting} and kernel compositions in \cite{duvenaud2013structure} are both related to our work in the sense that they search over a highly expressive generalization class in an unsupervised manner for models that explain observational data well.  In contrast we do full Bayesian inference, not greedy search, and the model family over which we search is ultimately more expressive as it is a high order language with stochastic primitives and, as a result, is capable of representing all computable probability distributions \cite{goodman2008church}.  
%
%\hspace{.5cm}
%
%This compositional approach to learning is not incompatible with work that seeks to find a good 
%representational language that could form a smoother expression space over which to search than the space of lambda expressions  \cite{liang2010learning}.
%
% Noah \cite{hwang2011inducing} whose main contribution might have been the idea of 
%generalization through searching the space from ``data regurgitation'' to simple procedure by using the implicit geometric prior on program length for its 
%Occam's razor effect.  This makes me think, what about biasing our inference towards programs that humans can read (by doing hierarchical modeling with 
%the pcfg's parameters as free variables and ``observed'' human-written program text).  These would give rise to awesome experiments where we show 
%performance vs. readability.  That would be super cool.
%
%The ideas and attempts to formalize the problem of automatic programming and automatic modelling via probabilistic programming have been introduced and described recently [Cite here Hwang, Vikash-Thesis, Venture, Andy Gordon, ...].
%

Our work relies heavily on a Turing-complete higher-order probabilistic programming language and system called Anglican \cite{wood2014anglican}, which borrows some of its modelling language syntax and semantics from Venture \cite{venture} and generally inherits  principles from Church \cite{goodman2008church}. What differentiates Anglican most substantially from the others is that it introduced and uses particle Markov Chain Monte Carlo \cite{andrieu2010particle} for probabilistic programming inference.
That we use Anglican means that we use PMCMC and Metropolis-Hastings algorithm for inference.

\section{Approach}
\label{sec:approach}
% !TEX root =  paper.tex

Our approach can be described in terms of a Markov Chain Monte Carlo (MCMC) approximate Bayesian computation (ABC) \cite{marjoram2003markov} targeting

\begin{equation}
\pi(\mathcal{X}|\hat{\mathcal{X}})p(\hat{\mathcal{X}}|\mathcal{T})p(\mathcal{T}),
\label{eqn:simpleabcformulation}
\end{equation} 

where at a high level $\pi(\mathcal{X}|\hat{\mathcal{X}})$ is a distance between summary statistics computed between observed data $\mathcal{X}$ and data, $\hat{\mathcal{X}}$, generated by interpreting latent sampler program text $\mathcal{T}$.

Consider first a single given data generating distribution $F$ with parameter vector $\theta$.  Let $\mathcal{X} = \{x_i\}_{i=1}^I, x_i \sim F(\cdot | \theta)$ be a set of samples from $F$.  Consider the task of learning program text $\mathcal{T}$ that when repeatedly interpreted returns samples whose distribution is close to $F$.   Let $\hat{\mathcal{X}}= \{\hat{x_j}\}_{j=1}^J$, $\hat{x_j} \sim \mathcal{T}(\cdot)$ be a set of samples generated by repeatedly interpreting $\mathcal{T}$ $J$ times..

Let $s$ be a summary function of a set of samples and let $d(s(\mathcal{X}),s(\hat{\mathcal{X}})) = \pi(\mathcal{X}|\hat{\mathcal{X}})$ be an unnormalized distribution function that returns high probability when $s(\mathcal{X}) \approx s(\hat{\mathcal{X}})$.  We refer to $d$ as a penalty, distance, or compatibility function interchangeably.

\begin{figure}[tbp]
\lstset{
  language=Lisp
}
\lstset{language=Lisp}
\begin{lstlisting}
[assume program-text (list `lambda `() (productions `() `real))]
[assume program (eval program-text)]
[assume samples (apply-n-times program 10000 '())]
[observe (normal (mean samples) noise-level) 0.0]
[observe (normal (variance samples) noise-level) 1.0]
[observe (normal (skewness samples) noise-level) 0.0]
[observe (normal (kurtosis samples) noise-level) 0.0]
[predict program-text]
\end{lstlisting}
\vspace{-4mm}
\caption{Probabilistic program to infer program text for a $\mathrm{Normal}(0, 1)$ sampler.}
\label{fig:inferring_std_normal_via_moments}
\end{figure}

We use probabilistic programming to write and perform inference in such a model, i.e. to generate samples of $\mathcal{T}$ from the marginal of \eqref{eqn:simpleabcformulation} and generalizations to come of the same.  The particular system we employ  \cite{wood2014anglican} uses PMCMC and MH for inference.
Refer to the probabilistic program code in Figure~\ref{fig:inferring_std_normal_via_moments} where the first line establishes a correspondence between $\mathcal{T}$and the variable \inline{program-text} then samples it from $p(\mathcal{T})$ where \inline{productions} is an adaptor-grammar-like \cite{johnson2007adaptor} prior on program text that is described in Section~\ref{sec:grammar}.
In this particular example $\theta$ is implicitly specified since the learning goal here is to find a sampler for the standard normal distribution. Also $\hat{\mathcal{X}}$ corresponds to the program variable \inline{samples} and, here, $J=10000$.   Here $s$ and $d$ are computed on the last four lines of the program with $s$ being implicitly defined as returning a four dimensional vector consisting of the estimated mean, variance, skewness, and kurtosis of the set of samples drawn from $\mathcal{T}$.  The distance function $d$ is also implicitly defined to be a multivariate normal  with mean $[0.0, 1.0, 0.0, 0.0]^T$ and diagonal covariance $\sigma^2\mathbf{I}$.  Note that this means that we are seeking sampler text whose output is distributed with mean 0, variance 1, skew 0, and kurtosis 0 and we penalize deviations from that by a squared exponential loss function with bandwidth $\sigma^2$, named \inline{noise-level} in the code .

This example highlights an important generalization of the original description of our approach.  For the standard normal example we chose a form of $s$ such that we can compute the first summary statistic of $d(s(F),s(\hat{\mathcal{X}}))$ analytically.   There are at least three kinds of scenarios in which $d$ can be computed in different ways.  The first occurs when we search for efficient code for sampling from known distributions.  In many such cases, as in the standard normal case just described, the summary statistics of $F$ can be computed analytically.  The second happens when we can only sample from $F$.    This corresponds to a situations when, for instance, there is a running, computationally expensive MCMC sampler that can be asked to produce additional samples.  This is how we frame compilation of probabilistic programs.    The third (how we originally described our approach) is the fixed dataset cardinality setting and corresponds to the setting of learning program text generative model for arbitrary observed data.

\begin{figure}[tbp]
\lstset{
  language=Lisp
}
\lstset{language=Lisp}
\begin{lstlisting}
[assume program-text (list `lambda `() (productions `(real) `bool))]
[assume program (eval program-text)]
[assume J 100]
[assume samples-1 (apply-n-times program J '(0.5))]
[observe (flip (G-test-p-value samples-1 `Bernoulli (list 0.5))) true]
$\ \ldots$
[assume samples-N (apply-n-times program J '(0.7))]
[observe (flip (G-test-p-value samples-N `Bernoulli (list 0.7))) true]
[predict program-text]
[predict (apply-n-times program J '(0.3))]
\end{lstlisting}
\vspace{-4mm}
\caption{Probabilistic program to infer program text for a $\mathrm{Bernoulli}(\theta)$ sampler and generate \inline{J} samples from the resulting procedure at a novel input argument value, \inline{0.3}.}
\label{fig:inferring_bernoulli_via_gtest}
\end{figure}

\begin{figure}[tbp]
\lstset{
  basicstyle=\tiny,
  language=Lisp
}
\begin{tabular}{ m{0.45\textwidth} m{0.45\textwidth} }
\begin{lstlisting}{}{program:abc}
[assume box-muller-normal
  (lambda (mean std)
    (+ mean (* std
      (* (cos (* 2 (* 3.14159
      (uniform-continuous 0.0 1.0))))
      (sqrt (* -2
        (log (uniform-continuous 0.0 1.0))))))))]
\end{lstlisting} & \begin{lstlisting}{}{program:abc}
[assume poisson (lambda (rate)
  (begin (define L (exp (* -1 rate)))
         (define inner-loop (lambda (k p)
         (if (< p L) (dec k)
           (begin (define u (uniform-continuous 0 1))
             (inner-loop (inc k) (* p u))))))
         (inner-loop 1 (uniform-continuous 0 1))))]
\end{lstlisting} \\
\end{tabular}
\vspace{-5mm}
\caption{Human-written sampling procedure program text for, left, $\textrm{Normal}(\mu, \sigma)$ \cite{box1958note} and, right, $\textrm{Poisson}(\lambda)$ \cite{knuth1998art}.  Counts of the constants, procedures, and expression expansions in these programs (and that of several other univariate samplers) are fed into our hierarchical generative prior over sampler program text.}
\label{fig:human_written_program_text}
\end{figure}

Figure~\ref{fig:inferring_bernoulli_via_gtest} illustrates the another important generalization of the formulation in \eqref{eqn:simpleabcformulation}.  When learning a standard normal sampler we did not have to take into account parameter values.  Interesting sampler program text is endowed with arguments, allowing it to generate samples from an entire family of parameterised distributions.   Consider the well known Box-Muller algorithm shown in Figure~\ref{fig:human_written_program_text}.  It is parameterized by  mean and standard deviation parameters.  For this reason we will refer to it and others like it as a conditional distribution samplers.  Learning conditional distribution sampler program text requires recasting our MCMC-ABC target slightly to include the parameter $\theta$ of the distribution $F$:

\begin{equation}
\pi(\mathcal{X}|\hat{\mathcal{X}},\theta)p(\hat{\mathcal{X}}|\mathcal{T}, \theta)p(\mathcal{T}|\theta)p(\theta).
\label{eqn:completeabcformulation}
\end{equation} 

Here in order to proceed we must begin to make approximating assumptions.  This is because in our case we need $p(\theta)$ to be truly improper as our learned sampler program text should work for all possible input arguments and not simply a just a high prior probability subset of values.  Assuming that program text that works for a few settings of input parameters is fairly likely to generalize well to other parameter settings we approximately marginalize our MCMC-ABC target \eqref{eqn:completeabcformulation} by choosing a small finite $N$ of $\theta_n$ parameters yielding our approximate marginalized MCMC-ABC target:

\begin{equation}
\frac{1}{N}\sum_{n=1}^N \pi(\mathcal{X}_n|\hat{\mathcal{X}}_n,\theta_n)p(\hat{\mathcal{X}}_n|\mathcal{T}, \theta_n)p(\mathcal{T}|\theta_n) \approx \int \pi(\mathcal{X}|\hat{\mathcal{X}},\theta)p(\hat{\mathcal{X}}|\mathcal{T}, \theta)p(\mathcal{T|\theta})p(\theta) d\theta.
\label{eqn:marginalizedabcformulation}
\end{equation} 

The probabilistic program for learning conditional sampler program text for $\mathrm{Bernoulli}(\theta)$ in Figure~\ref{fig:inferring_bernoulli_via_gtest} shows an example of this kind of approximation.  It samples from $\mathcal{T}$ $N$ times, accumulating summary statistic penalties for each invocation.   In this case each individual summary distance computation involves computing both a G-test statistic 
\[G_n= 2 \sum_{i\in{0,1}} \#[\hat{\mathcal{X}}_n = i]\mathrm{ln}\left(\frac{\#[\hat{\mathcal{X}}_n = i]}{\theta_n^i(1-\theta_n)^{(1-i)} \cdot |\hat{\mathcal{X}}_n|}\right),\]

where  $\#[\hat{\mathcal{X}}_n = i]$ is the number of samples in $\hat{\mathcal{X}}_n$ that take value $i$ and its corresponding p-value under the null hypothesis that $\hat{\mathcal{X}}_n$ are samples from $\mathrm{Bernoulli}(\theta_n)$.  Since the G-test statistic is approximately $\chi^2$ distributed, i.e.~$G\sim\chi^2(1)$, we can construct $d$ in this case by computing the probability of falsely rejecting the null hypothesis  $H_0 : \hat{\mathcal{X}}_n \sim \mathrm{Bernoulli}(\theta_n)$.  Falsely rejecting a null hypothesis is equivalent to flipping a coin with probability given by the p-value of the test and having it turn up heads.  These are the summary statistic penalties accumulated in the \inline{observe} lines in Figure~\ref{fig:inferring_bernoulli_via_gtest}. 

 As an aside, in the probabilistic programming compilation context $\theta$ could be all of the \inline{observe}'d data in the original program.  By this parameterising compilation links our approach to that of \cite{hwang2011inducing}.

\subsection{Grammar and production rules}
\label{sec:grammar}
As we have the expressive power of a higher-order probabilistic programming language at our disposal, our prior over conditional distribution sampler program text is quite expressive.  At a high level it is similar to the adaptor grammar \cite{johnson2007adaptor} prior used in \cite{liang2010learning} but diverges in details, particularly those having to do with creation of local environments and the conditioning of subexpression choices on type signatures. In pseudocode our prior can be expressed as follows:
\begin{enumerate}[leftmargin=*]
\item $expr_{type} | env \rightarrow$ a random variable name from the $env$ with type $type$.
\item $expr_{type} | env \rightarrow$ a random constant with the type $type$.
 Constants with types integer, real, etc. are sampled from a Chinese restaurant process (CRP) representation of a marginalized discrete Dirichlet process prior pair with $DP_{type}(H_{type}, \alpha)$ for each type, where the base distribution $H_{type}$ is itself a mixture distribution. For example for type real we use a mixture of \inline{(normal 0 10)}, \inline {(uniform-continuous -100 100)}, and  uniform over common constants like $\{-1, 0, 1, \ldots\}$.
\item $expr_{type} | env \rightarrow (procedure_{type}\ expr_{arg\_1\_type}\ ...\  expr_{arg\_N\_type})$, where $procedure$ is a primitive or stochastic procedure in the global environment with output type signature $type$ sampled randomly.
\item $expr_{type} | env \rightarrow (compound\_procedure_{type}\ expr_{arg\_1\_type}\ ...\  expr_{arg\_N\_type})$, where {\tt $compound\_procedure_{type}$} is a compound procedure sampled from a CRP representation of a marginalized discrete Dirichlet process prior pair with $DP_{type}(G_{type}, \beta)$. The base distribution $G_{type}$ generates compound procedures with return type $type$, Poisson distributed argument count, and random parameter types. The body of the compound procedure is  generated using the same production rules given an environment that incorporates argument input variable names and values.
\item $expr_{type} | env \rightarrow (let\ [(gensym)\ expr_{real} ]\ expr_{type} | env \cup gensym))$, where $ env \cup gensym$ is an extended environment with the variable named $(gensym)$ and its value added).
\item $expr_{type} | env \rightarrow (if\ (expr_{bool})\ expr_{type}\ expr_{type})$.
\item $expr_{type} | env \rightarrow (recur\ expr_{arg\_1\_type}\ ...\  expr_{arg\_M\_type})$, i.e. recursive call to the current compound procedure.
\end{enumerate}

To avoid numerical errors while interpreting generated programs we replace functions like \inline{log(a)} with \inline{safe-log(a)}, which returns $0$ if $a < 0$, and \inline{uniform-continuous}  with \inline{safe-uc(a, b)} which swaps arguments if $a > b$ and returns $a$ if $a = b$. 

The set of types we used for our experiments was \texttt{\{real, bool\}} and the general set of procedures in the global environment included \texttt{+, $-$, *, safe-div, safe-uc}.  

\subsection{Production rule probabilities}

While it is possible to manually specify production rule probabilities for the grammar in Section~\ref{sec:grammar} we took a hierarchical Bayesian approach instead, learning from human-written sampler source code.  To do this we translated existing implementations of common one-dimensional statistical distribution samplers \cite{Devroye86non-uniformrandom} into Anglican source. Examples are provided in Figure~\ref{fig:human_written_program_text}.  Conveniently all of them require only one stochastic procedure \inline{uniform-continuous} so we also only include that single stochastic procedure in our grammar.   

We compute held-out production rules prior probabilities from this corpus in cross-validation way so that when we are inferring a probabilistic program to sample from $F$ we update our priors using counts from all other sampling code in the corpus, specifically excluding the sampler we are attempting to learn.   Our production rule probability estimates are smoothed by Dirichlet priors. Note that in the following experiments (Sections~\ref{sec:experiments_generalizing_empirical_data} and~\ref{sec:experiments_compilation}) the production rule priors were updated then fixed during inference. True hierarchical coupling and joint inferences approaches are straightforward from a probabilistic programming perspective \cite{maddison2014structured} but result in inference runs that take longer to compute.

\section{Experiments}
\label{sec:experiments}
% !TEX root =  paper.tex

The experiments we perform illustrate all three uses cases outlined for automatically learning probabilistic programs.  We begin by illustrating the 
expressiveness of our prior over sampler program text in Section~\ref{sec:experiments_samples_from_sampled_programs_from_prior}. We then report results from experiments in which we test our approach in all three scenarios for how we can compute the ABC penalty $d$. The first set of experiments in Section~\ref{sec:experiments_learning_one_dimensional_distributions}) tests our ability to learn probabilistic programs that produce samples from known one-dimensional probability distributions.  In these experiments $d$ either probabilistically conditions on $p$-values of one-sample statistical hypothesis tests or on approximate moment matching. The second set of experiments in Section~\ref{sec:experiments_generalizing_empirical_data}) addresses the cases where only a finite number of samples from an unknown real-world source are provided. The final experiment in Section~\ref{sec:experiments_compilation}) is a preliminary study in probabilistic program compilation where it is possible to gather a continuing set of samples.

\subsection{Samples from sampled probabilistic programs}
\label{sec:experiments_samples_from_sampled_programs_from_prior}

\begin{figure}[tbp]
\includegraphics[width=0.33\textwidth]{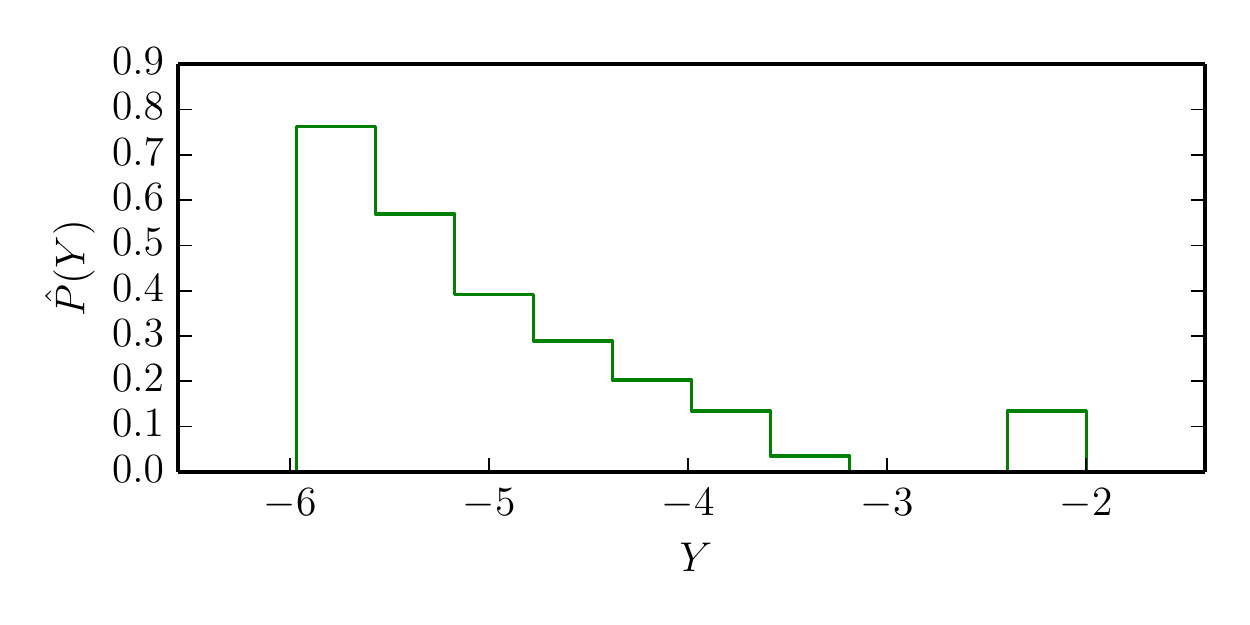}
\includegraphics[width=0.33\textwidth]{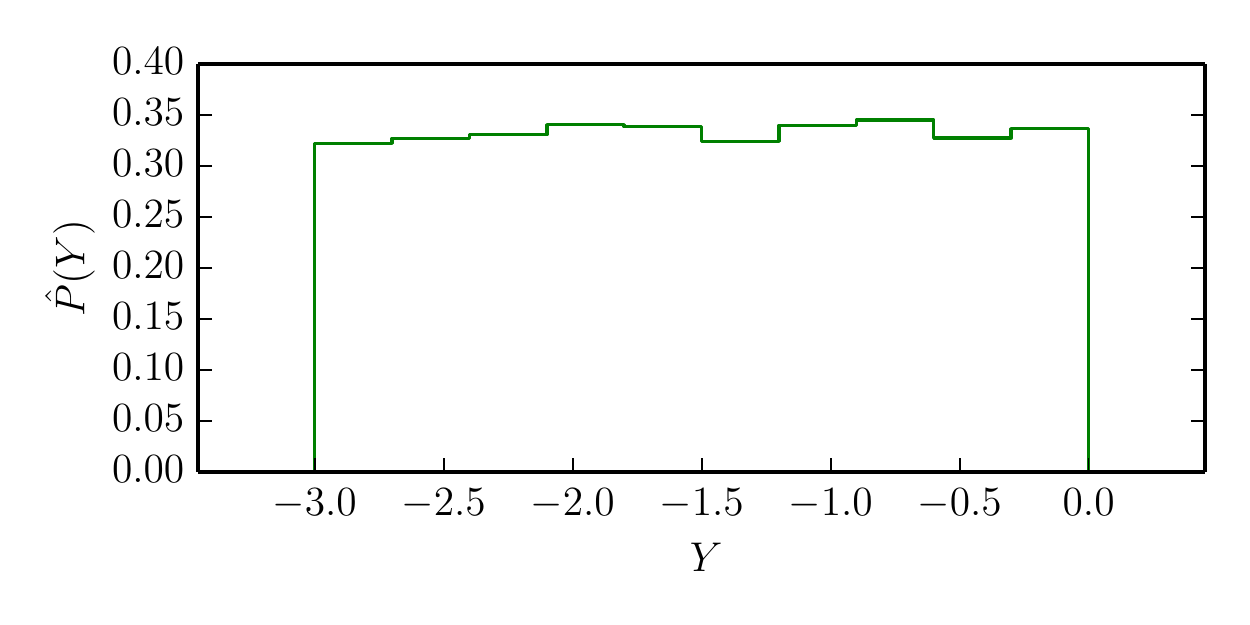}
\includegraphics[width=0.33\textwidth]{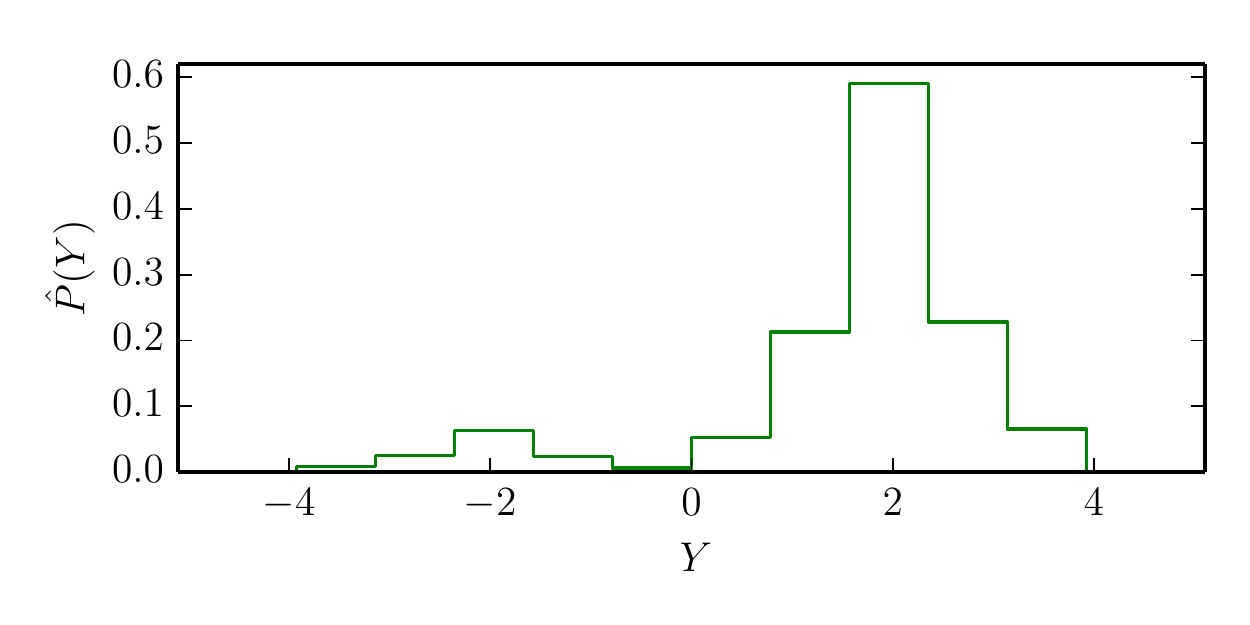}
\includegraphics[width=0.33\textwidth]{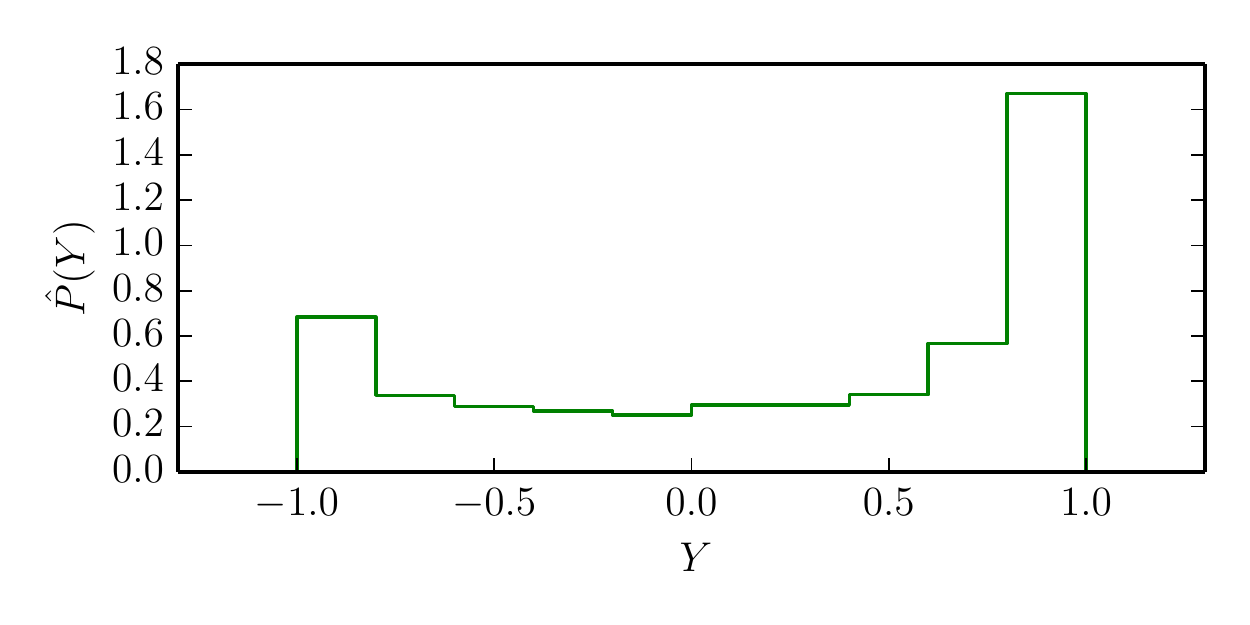}
\includegraphics[width=0.33\textwidth]{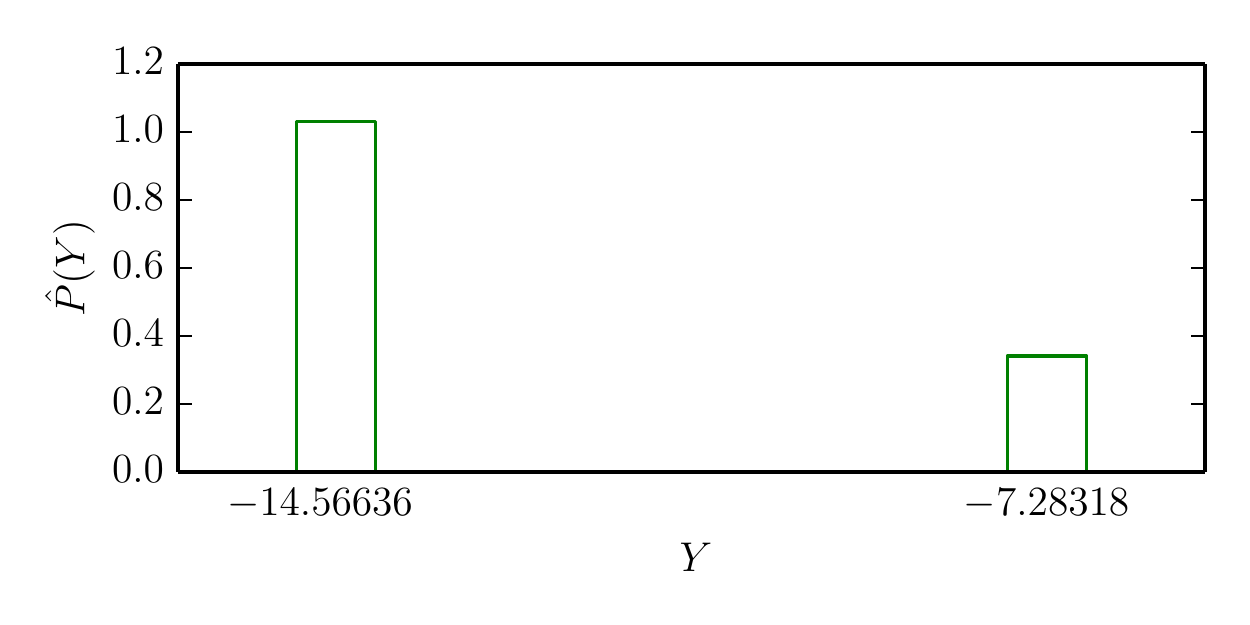}
\includegraphics[width=0.33\textwidth]{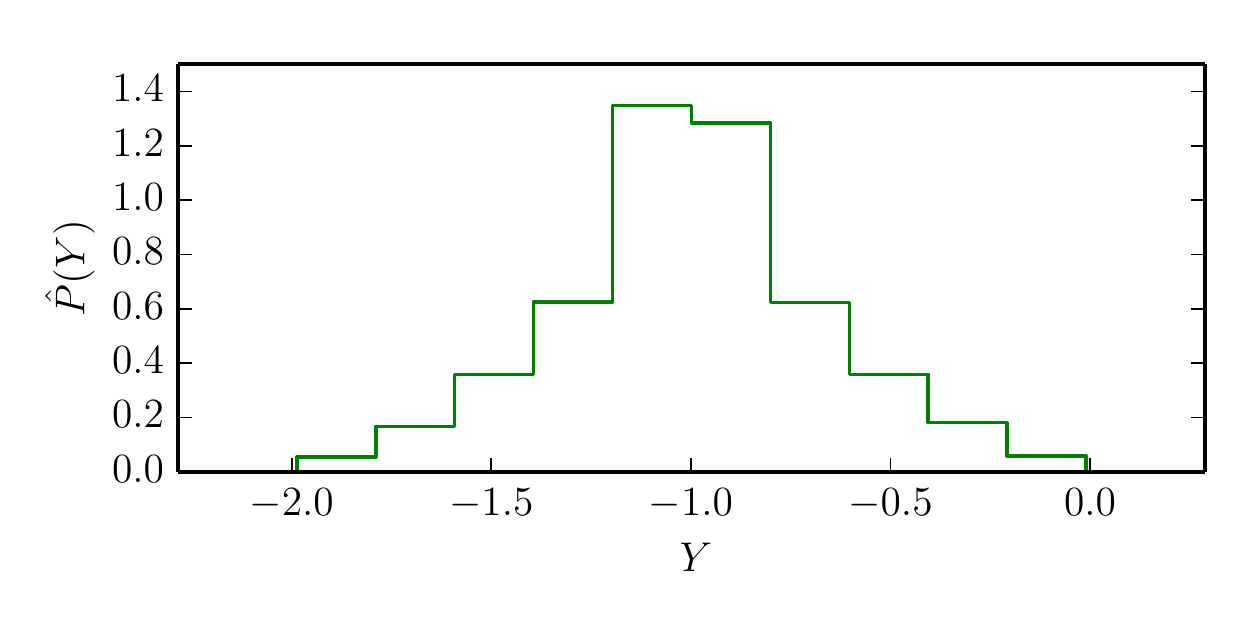}
\caption{Histograms of samples generated by repeatedly evaluating probabilistic procedures sampled from our prior over probabilistic sampling procedure text.  The prior is constrained to generate samplers with univariate output but is clearly otherwise  flexible enough to represent a nontrivial spectrum of distributions.}
\label{fig:samples_from_probabilistic_programs_from_prior}
\end{figure}

To illustrate the flexibility of our prior, specifically the production rules we employ, we show samples generated by probabilistic programs sampled from the prior in Section~\ref{sec:grammar}.   In Figure~\ref{fig:samples_from_probabilistic_programs_from_prior} we show six histograms of samples from six sampled probabilistic programs from our prior over probabilistic programs. Such randomly generated samplers constructively define considerably different distributions.  Note in particular the variability of the domain, variance, and even number of modes.  

\begin{figure}[tb]
\includegraphics[width=0.33\textwidth]{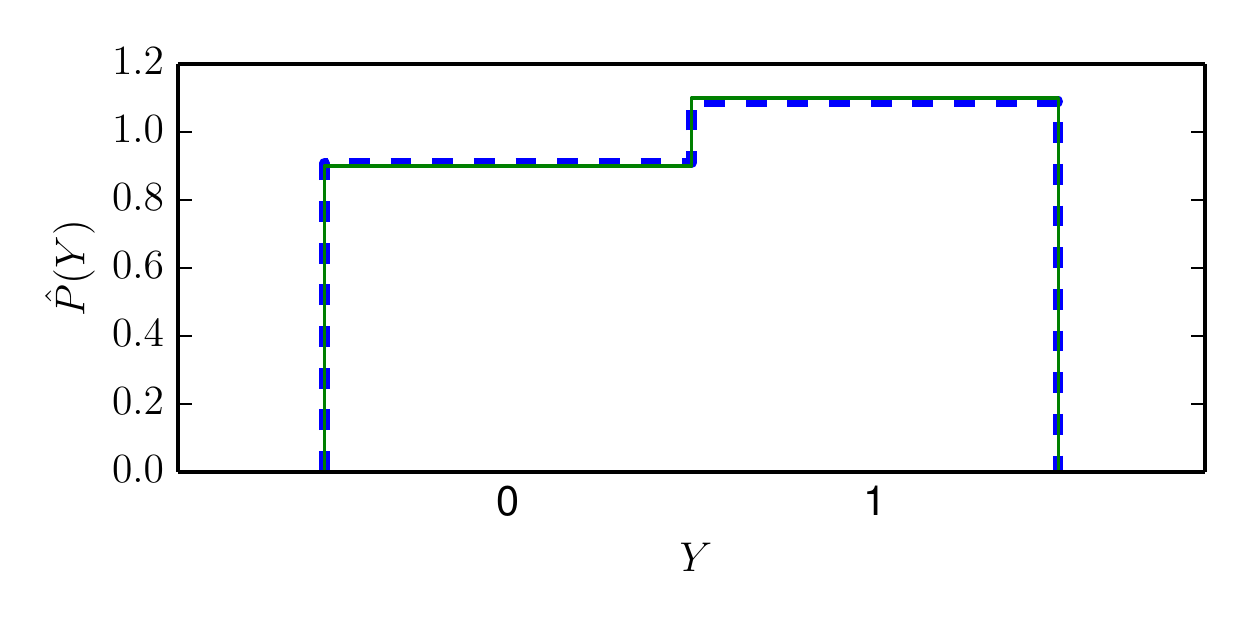}
\includegraphics[width=0.33\textwidth]{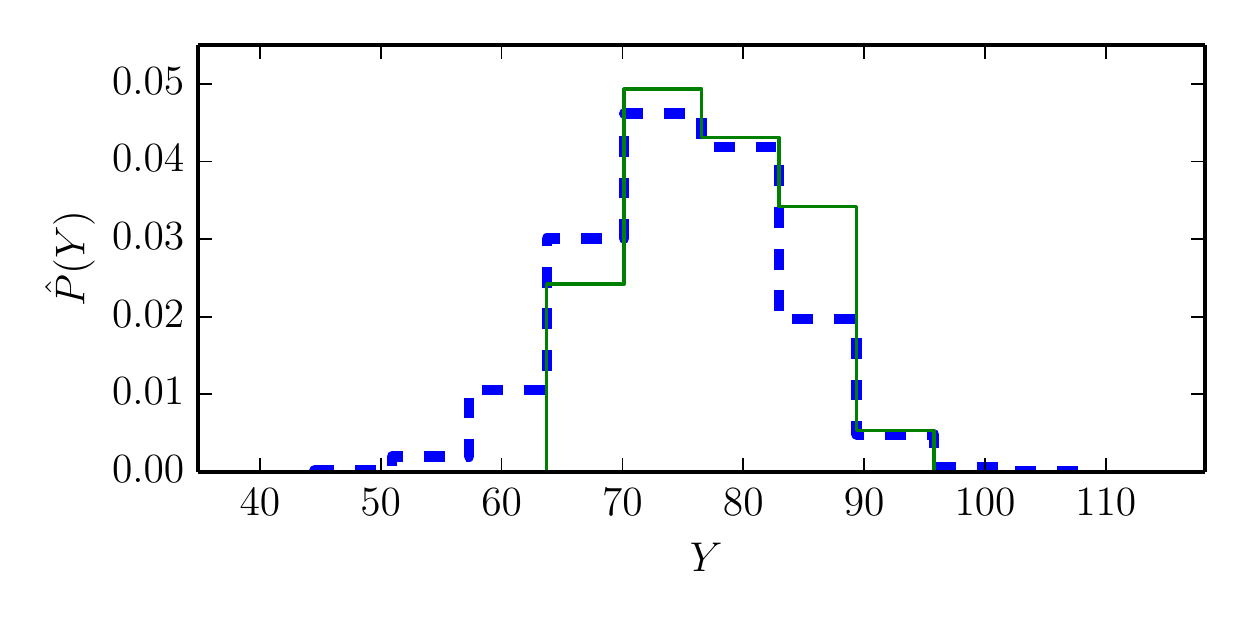}
\includegraphics[width=0.33\textwidth]{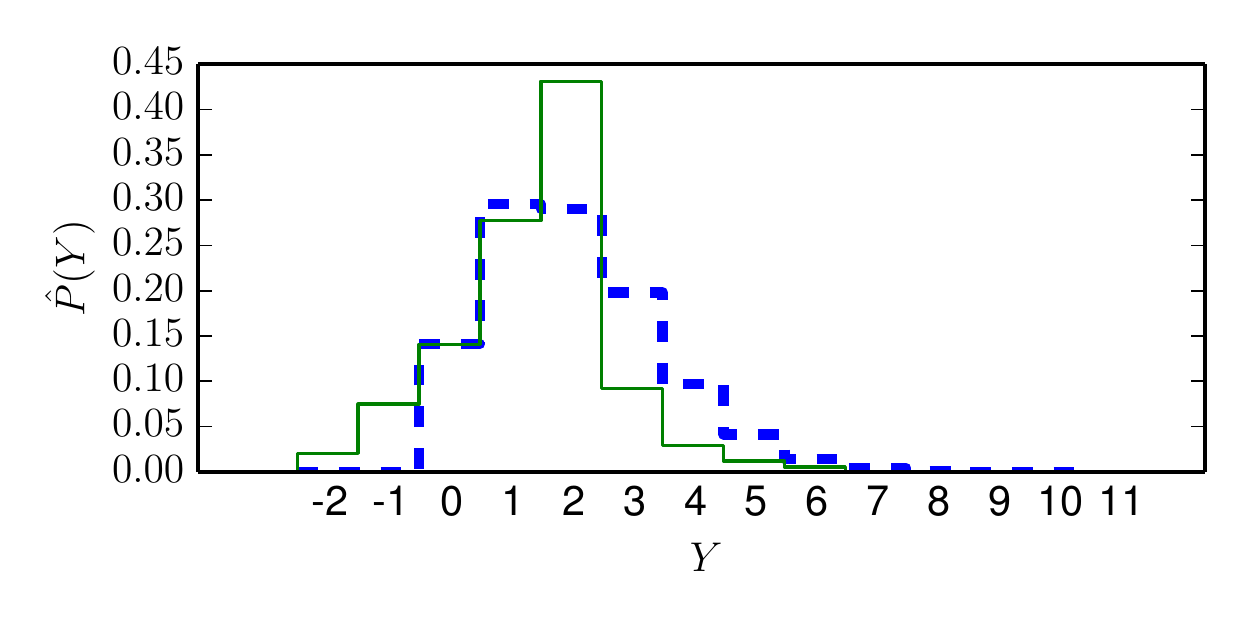}
\includegraphics[width=0.33\textwidth]{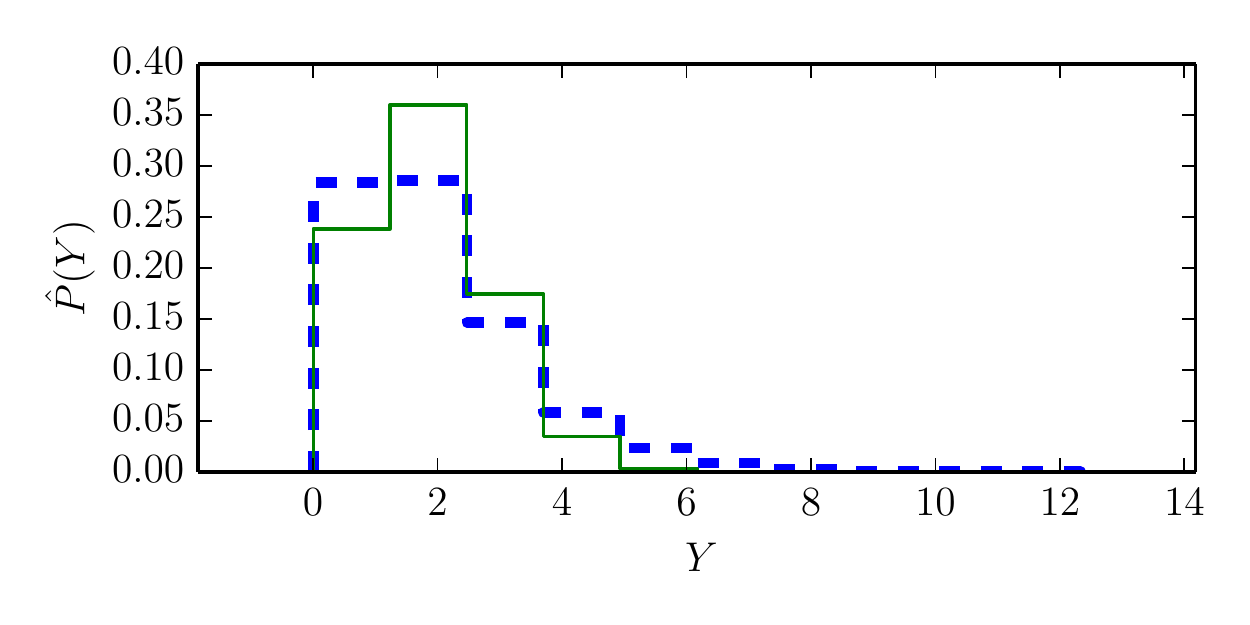}
\includegraphics[width=0.33\textwidth]{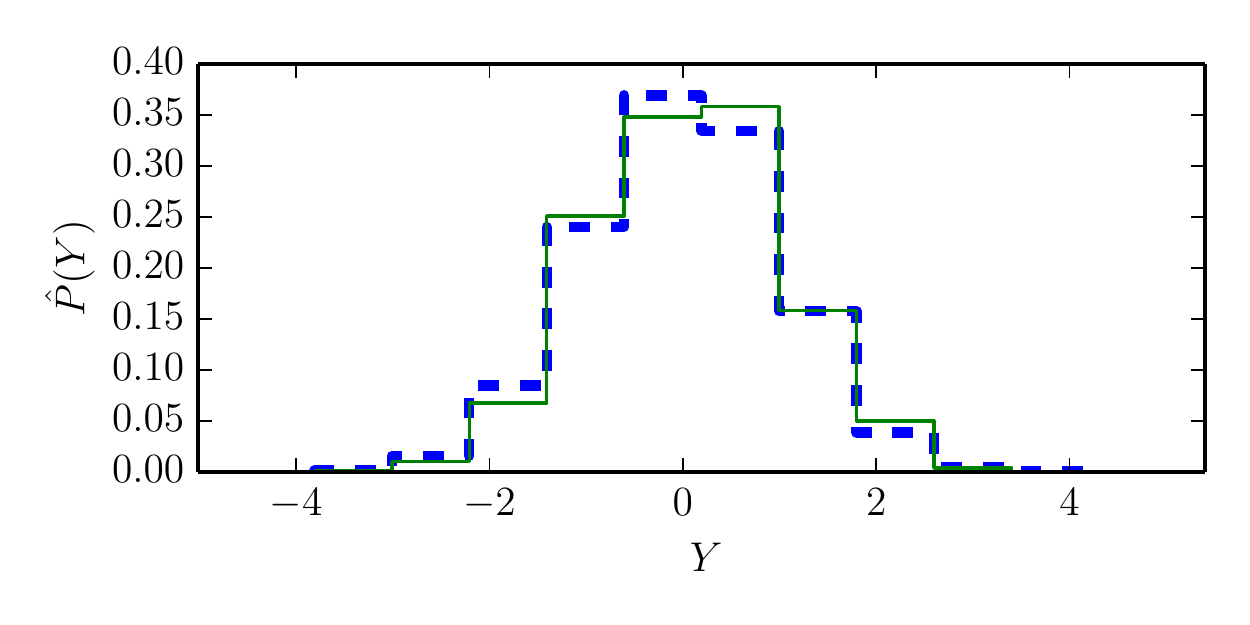}
\includegraphics[width=0.33\textwidth]{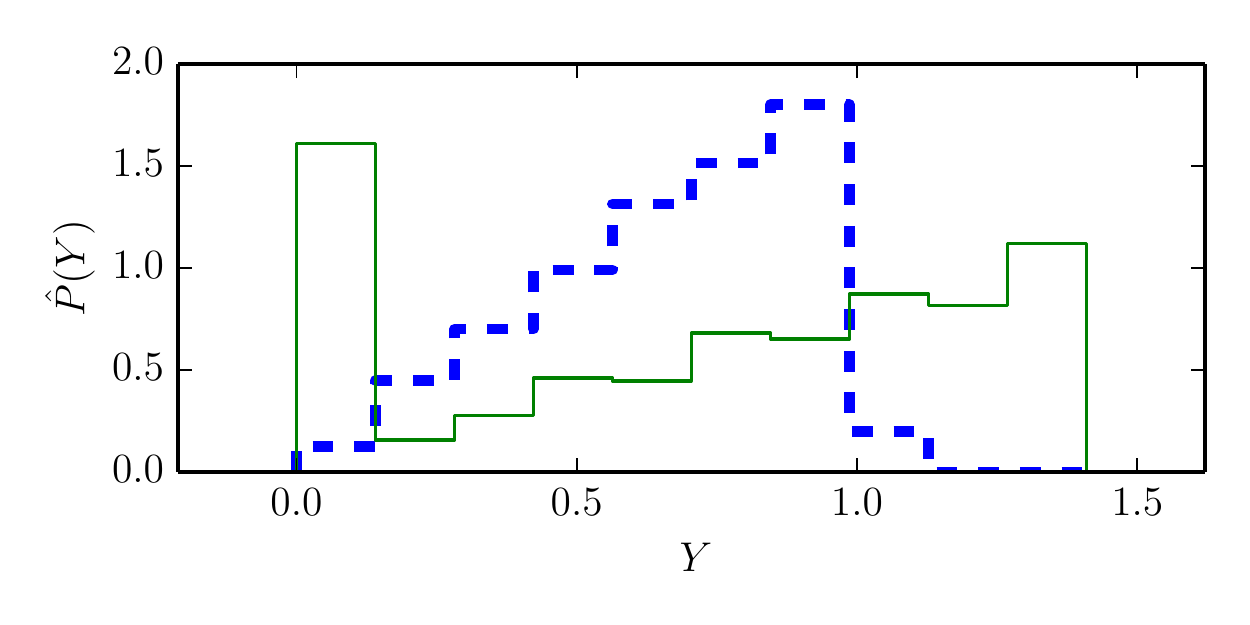}
\caption{Representative histograms of samples {\it (green solid lines)} drawn by repeatedly interpreting inferred sampler program text versus {\it (blue dashed lines)} histograms of exact samples drawn from the corresponding true distribution.  Top row left to right: $\mathrm{Bernoulli}(p)$, $\mathrm{Normal}(\mu, \sigma)$, $\mathrm{Poisson}(\lambda)$.  Bottom row same:
 $\mathrm{Gamma}(a, 1.0)$,
$\mathrm{Normal}(0, 1)$,  
$\mathrm{Beta}(a, 1.0)$.
The parameters used to produce these plots do not appear in the training data.  In the case of  $\mathrm{Bernoulli}(p)$ we inferred programs that sample exactly from the true distribution (see Figure~\ref{program:Bernoulli_samplers}).  Not all finite-time inference converges to good approximate sampler code as illustrated by the $\mathrm{Beta}(a, 1.0)$ example.
}
\label{fig:common_distributions_histograms}
\end{figure}

\begin{figure}
\lstset{
  basicstyle=\tiny,
  language=Lisp
}
\begin{tabular}{ m{0.55\textwidth} m{0.40\textwidth} }
\begin{lstlisting}{}{program:abc}
(lambda (par stack-id) (* (begin (define sym0 0.0)
    (exp (safe-uc -1.0 (safe-sqrt (safe-uc
    (safe-div (safe-uc 0.0 (safe-uc 0.0 3.14159))
      par) (+ 1.0 (safe-uc (begin (define sym2
      (lambda (var1 var2 stack-id) (dec var2)))
      (sym2 (safe-uc -2.0 (* (safe-uc 0.0 (begin
      (define sym4 (safe-uc sym0 (* (+ (begin
      (define sym5 (lambda (var1 var2 stack-id)
      (safe-div (+ (safe-log (dec 0.0)) -1.0) var1)))
      (sym5 (exp par) 1.0 0)) 1.0) 1.0))) (if (< (safe-uc
      par sym4) 1.0) sym0 (safe-uc 0.0 -1.0)))) sym0))
      (safe-div sym0 (exp 1.0)) 0)) 0.0))))))) par))
\end{lstlisting} & \begin{lstlisting}{}{program:abc}
(lambda (stack-id)
  (* 2.0 (* (*
      (* -1.0 (safe-uc 0.0 2.0))
      (safe-uc (safe-uc 4.0
        (+ (safe-log 2.0) -1.0))
        (* (safe-div 2.0
          -55.61617747203855)
          (if (< (safe-uc
            (safe-uc
            27.396810474207317
             (safe-uc -1.0 2.0)) 2.0) 2.0)
            4.0 -1.0)))) -1.0)))
\end{lstlisting} \\
\end{tabular}
\vspace{-7mm}
\caption{Inferred univariate sampler program text for {\it (left)} $\mathrm{Gamma}(a, 1)$ and {\it (right)} observational data of unknown distribution.  These two program respectively generated the green histograms on the bottom left of Figure~\ref{fig:common_distributions_histograms} and the right of Figure~\ref{fig:real_world_data_approximation_histograms}.
These programs were manually simplified for display, i.e. substitutions like \inline{0.0} for \inline{(* 1.0 0.0)} were performed.}
\label{fig:found_approximators_as_probabilistic_programs}
\end{figure}

\subsection{Learning sampler code for common one-dimensional distributions}
\label{sec:experiments_learning_one_dimensional_distributions}

Source code exists for efficiently sampling from many if not all common one-dimensional distributions.  We conducted experiments to test our ability to automatically discover such sampling procedures and found encouraging results.

In particular we performed a set of leave-one-out styles experiments to infer sampler program text for six common one-dimensional distributions: $\mathrm{Bernoulli}(p)$, $\mathrm{Poisson}(\lambda)$, $\mathrm{Gamma}(a, 1.0)$, $\mathrm{Beta}(a, 1)$, $\mathrm{Normal}(0, 1)$, $\mathrm{Normal}(\mu, \sigma)$.
For each distribution we performed MCMC-ABC inference with approximately marginalizing over the parameter space using a small random set $\left\{\theta_1, \ldots, \theta_N\right\}$ of parameters and conditioning on statistical hypothesis tests or on moment matching as appropriate.  Note that the pre-training of the hierarchical program text prior was never given the text of the sampler for the distribution being learned.

Representative histograms of samples from the best posterior program text sample discovered in terms of summary statistics match are shown in Figure~\ref{fig:common_distributions_histograms}.
A pleasing result is the discovery of the exact $\mathrm{Bernoulli}(p)$ distribution sampler program, the text of which is shown in Figure~\ref{program:Bernoulli_samplers}.  Figure~\ref{fig:found_approximators_as_probabilistic_programs} shows the inferred sampler text for $\mathrm{Gamma}(a, 1)$.  How to fully characterize the divergence between learned sampling algorithms and the true distribution via a mechanisms other than exhaustive computation and hypothesis testing remains an open question. 

\begin{figure}[tb]
\lstset{
  language=Lisp
}
\begin{lstlisting}
(lambda (par stack-id) (if (< (uniform-continuous 0.0 1.0) par) 1.0 0.0))

(lambda (par stack-id)
  (if (< 1.0 (safe-sqrt (safe-div par (safe-uc par (dec par))))) 1.0 0.0))
(lambda (par stack-id)
  (if (< 1.0 (safe-uc (safe-sqrt par) (+ par (cos par)))) 1.0 0.0))
\end{lstlisting}
\vspace{-4mm}
\caption{{\it (top)} Human-written exact $\mathrm{Bernoulli}(p)$ sampler. {\it (bottom two)} Inferred sampler program text.  The first is also an exact sampler for $\mathrm{Bernoulli}(p)$. The last is another sampler also assigned non-zero posterior probability but it is not exact.}
\label{program:Bernoulli_samplers}
\end{figure}

\subsection{Generalizing arbitrary data distributions}
\label{sec:experiments_generalizing_empirical_data}

\begin{figure}
\vspace{-2mm}
\includegraphics[width=0.33\textwidth]{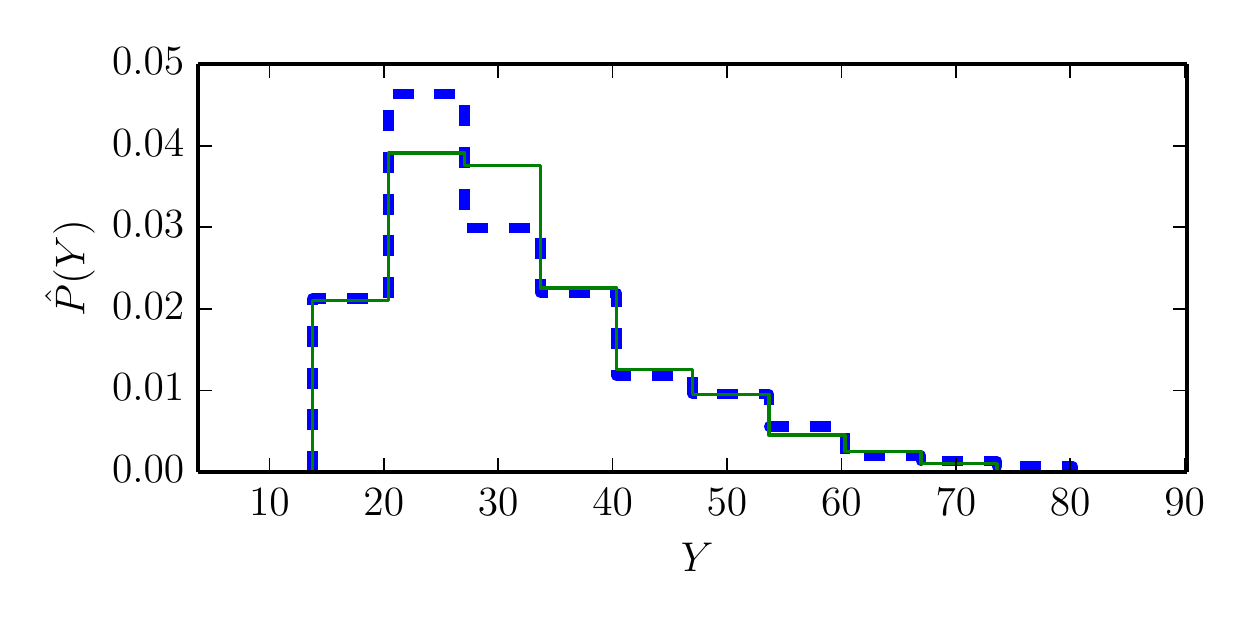}
\includegraphics[width=0.33\textwidth]{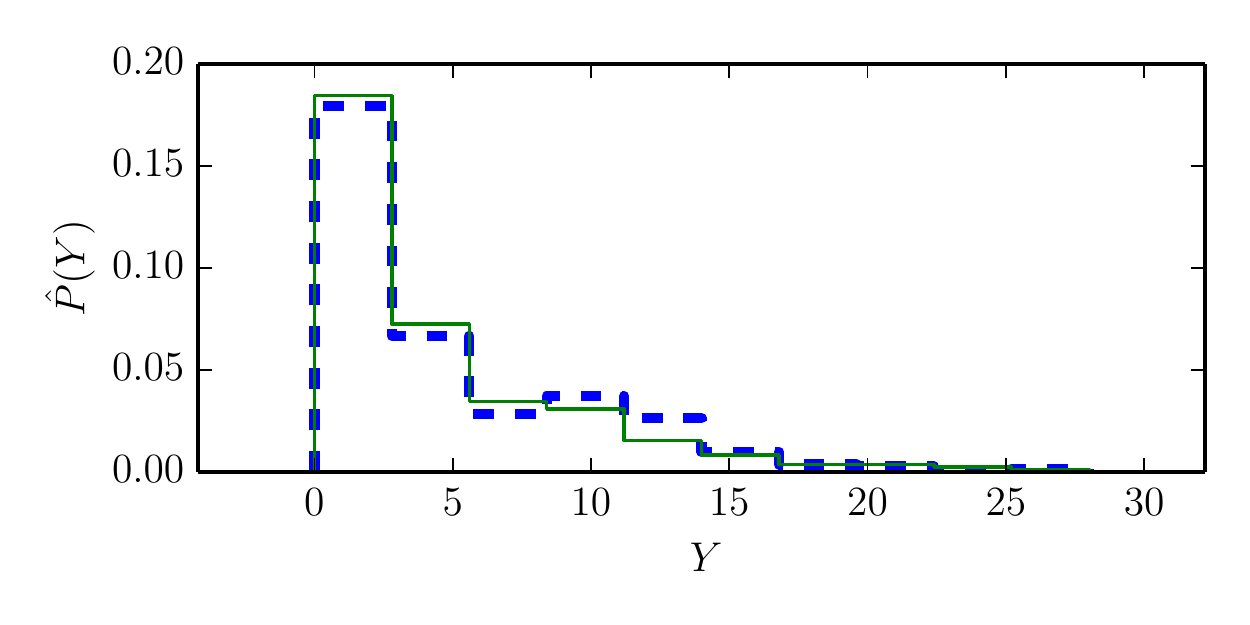}
\includegraphics[width=0.33\textwidth]{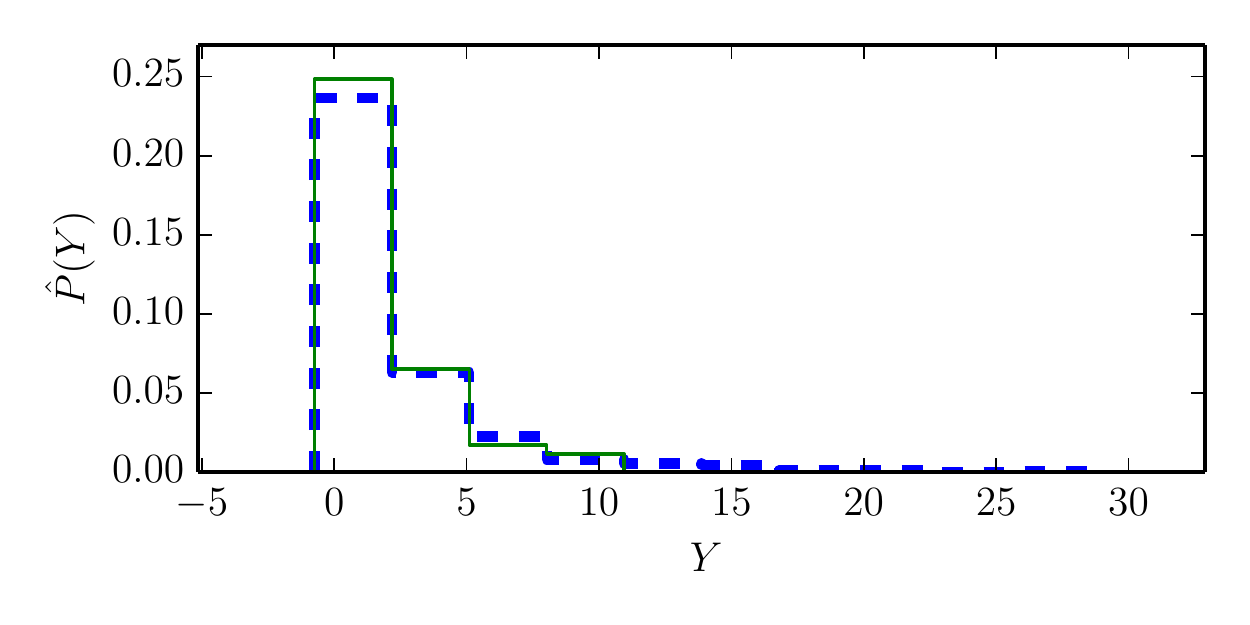}
\caption{Histograms of samples {\it (green solid lines)} generated by repeatedly interpreting inferred sampler program text and the empirical distributions {\it (blue dashed)} which they were trained to match.}
\label{fig:real_world_data_approximation_histograms}
\end{figure}

We also explored using our approach to learn generative models in the form of sampler program text for real world data of unknown distribution. We arbitrarily chose three continuous indicator features from a credit approval dataset \cite{quinlan1987simplifying,Bache+Lichman:2013} and inferred sampler program text using two-sample Kolmogorov-Smirnov distribution equality tests (vs. the empirical data distribution) analogously to G-test described before. Histograms of samples from the best inferred sampler program text versus the training empirical distributions are shown in Figure~\ref{fig:real_world_data_approximation_histograms}. An example inferred program is shown in Figure~\ref{fig:found_approximators_as_probabilistic_programs}~{\it (right)}.  The data distribution representation, despite being expressed in the form of sampler program text, matches salient characteristics of the empirical distribution well.  

\subsection{Compilation of probabilistic programs}
\label{sec:experiments_compilation}

MCMC sampling, particularly in a Bayesian context, is usually quite costly and further, requires large amounts of storage to represent the learned distribution as samples.   Learning an representation of the posterior in terms of a sampling procedure that directly samples approximately from the posterior distribution of interest could potentially improve both, particularly for the purposes of repeated posterior predictive inference.
In probabilistic programming where sample-based posterior representations are the only option the problem is particularly acute.  Further, higher-order probabilistic programming languages require that the expressivity class of the approximating distribution is at least the same as that of another probabilistic program. While the ultimate aim of compilation of probabilistic program inference to a learned program for sampling directly from the posterior of interest remains quite far off, our preliminary experiments are encouraging.  

To explore this possibility we took an uncollapsed Beta-Binomial model with prior distribution $\theta \sim \mathrm{Beta}(1.0, 1.0)$, and used Metropolis Hastiings to infer a sampled-based representation of the posterior distribution over $\theta$ given four successful trials from $\mathrm{Bernoulli}(\theta)$. The correspondent probabilistic program is given in Figure~\ref{program:CompilationCase}.
Then we used our approach to learn a probabilistic program that when repeatedly invoked produces samples statistically match the empirical posterior distribution.  Examples of inferred probabilistic procedures are given in Figure~\ref{program:CompilationCase} {\it (right)}. The analytical posterior distribution in this case is $\mathrm{Beta}(5.0, 1.0)$ to which we found good approximations. Note that in the probabilistic program compilation experiment additional primitives include \inline{beta}, \inline{normal}, and other higher order stochastic procedures were added to the program text generative model.

\begin{figure}[tb]
\lstset{
  basicstyle=\scriptsize,
  language=Lisp
}
\begin{tabular}{ m{0.33\textwidth} m{0.40\textwidth} }
\begin{lstlisting}{}{program:abc}
[assume theta (beta 1.0 1.0)]
[observe (flip theta) True]
[observe (flip theta) True]
[observe (flip theta) True]
[observe (flip theta) True]
[predict theta]
\end{lstlisting} & \begin{lstlisting}{}{program:abc}
[assume theta (safe-beta 4.440 1.0)]
[assume theta (safe-sqrt (safe-beta (safe-log 11.602) 1.0))]
[assume theta (safe-beta (safe-sqrt 27.810) 1.0)]

[assume theta (beta 5.0 1.0)]
[predict theta]
\end{lstlisting} \\
\end{tabular}
\vspace{-6mm}
\caption{ {\it (left)} Uncollapsed Beta-Binomial model as a probabilistic program. We are interested in the posterior distribution over the latent variable $\theta$. {\it (right, top)} The salient line from three inferred probabilistic programs (i.e. the result of probabilistic programs compilation) which produce samples of $\theta$ that are statistically similar in distribution to the posterior distribution induced by the original probabilistic program. Each complete program ended with the line \inline{[predict theta]}.
{\it (right, bottom)} Human-written code for exactly sampling from the analytical posterior.  The inferred compiled posterior samplers are indeed close the exact sampler.}
\label{program:CompilationCase}
\end{figure}

\section{Discussion}
\label{sec:discussion}
% !TEX root =  paper.tex

Our novel approach to program synthesis via probabilistic programming raises at least as many questions as it answers.  One key high level question this kind of work sharpens is, really, what is the goal of program synthesis?  By framing program synthesis as a probabilistic inference problem we are implicitly naming our goal to be that of estimating a distribution over programs that obey some constraints rather than as a search for a single best program that does the same. On one hand, the notion of regularising via a generative model is natural as doing so predisposes inference towards discovery of programs that preferentially possess characteristics of interest (length, readability, etc.).  On the other hand,  exhaustive computational inversion of a generative model that includes evaluation of program text will clearly remain intractable for the foreseeable future.  For this reason greedy and stochastic search inference strategies are basically the only options available.  We employ the latter, and MCMC in particular, to explore the posterior distribution of programs whose outputs match constraints knowing full-well that its actual effect in this problem domain, and, in particular finite time, is more-or-less that of stochastic search. We could add an annealing temperature and schedule \cite{van1987simulated} to clarify our use of MCMC as search, however, while ergodic, our system is sufficiently stiff to not require quenching (and as a result almost certainly will not achieve maxima in general).

It is  pleasantly surprising, however, that the Monte Carlo techniques we use were able to find exemplar programs in the posterior distribution that actually do a good job of generalising observed data in the experiments we report.  It remains an open question whether or not sampling procedures are the best stochastic search technique to use for this problem in general however.  Perhaps by directly framing the problem as one of search we might do better, particularly if our goal is a single best program. Techniques ranging from genetic algorithms \cite{poli2008field} to Monte Carlo tree search \cite{browne2012survey} all show promise and bear consideration.

One interesting way to take this work forward is to introduce techniques from the cumulative/incremental learning  community \citep{dechter2013bootstrap}, perhaps by adding time-dependent and hierarchical dimensions to the program text generative model.  In the specific context of learning sampler program text, it would be convenient if, for instance when learning the program text for sampling from a parameterised normal distribution, one had access to an already learned subroutine for sampling from a standard normal.  In related work from the field of inductive programming large gains in performance were observed when the learning task was structured in this way \citep{henderson2010incremental}.  

Our example inference tasks are just the start.  What inspired and continues to inspire us is our the internal experience of our own ability to reason about procedure.   Given examples, humans clearly are able to generate program text for procedures that compute or otherwise match examples.  Humans can  physically  simulate Turing machines, and, it would seem clear, are capable doing something at least as powerful when deducing the action of a particular piece of program text from the text itself.  No candidate artificial intelligence solution will be complete without the inclusion of such ability.  Those without will always be deficient in the sense that it is apparent that humans can internally represent and reason about procedure.  Perhaps some generalised representation of procedure is the actual expressivity class of human reasoning.  It certainly can't be less.

\subsubsection*{Acknowledgments}

The authors thank many people for their help, wholesome discussions, suggestions, and comments, including Brooks Paige, Jan-Willem van de Meent, David Tolpin and Tejas Kulkarni. This work was supported by Xerox faculty research award and Somerville college scholarship. Any opinions, findings, and conclusions or recommendations expressed in this work are those of the authors and do not necessarily reflect the views of any of the above sponsors.

This material is based on research sponsored by DARPA through the U.S. Air Force Research Laboratory under Cooperative Agreement number FA8750-14-2-0004.  The U.S. Government is authorized to reproduce and distribute reprints for Governmental purposes notwithstanding any copyright notation heron. The views and conclusions contained herein are those of the authors and should be not interpreted as necessarily representing the official policies or endorsements, either expressed or implied, of DARPA, the U.S. Air Force Research Laboratory of the U.S. Government.

\subsubsection*{References}
\begingroup
\renewcommand{\section}[2]{}%
\bibliographystyle{unsrtnat}
\small
\bibliography{references}
\endgroup

\end{document}